%% file: main.tex
\documentclass[runningheads]{llncs}

\usepackage{eccv}

\usepackage{eccvabbrv}

\usepackage{graphicx}
\usepackage{amsmath}
\usepackage{amssymb}
\usepackage{booktabs}
\usepackage{bm}
\usepackage{amsfonts}
\usepackage{xcolor}
\usepackage{multirow}
\usepackage{float}

\usepackage[accsupp]{axessibility}  

\usepackage{hyperref}

\usepackage{orcidlink}

\usepackage[capitalize]{cleveref}
\crefname{section}{Sec.}{Secs.}
\Crefname{section}{Section}{Sections}
\Crefname{table}{Table}{Tables}
\crefname{table}{Tab.}{Tabs.}

\newcommand{\ourdataset}[1]{WhatifVideo-1.0}
\newcommand{\ourmethod}[1]{ReimaginedAct}
\begin{document}

\title{Action Reimagined:  Text-to-Pose Video Editing for Dynamic Human Actions} 


\author{Lan Wang\inst{1} \and
Vishnu Boddeti \inst{1} \and
Ser Nam Lim\inst{2} }

\authorrunning{Lan Wang et al.}

\institute{\textsuperscript{1}Michigan State University \textsuperscript{2}University of Central Florida\\
\email{\{wanglan3,vishnu\}@msu.edu}
\email{sernam@ucf.edu}}

\maketitle

\input{00_abstract}

\input{01_introduction}
\input{02_relatedwork}
\input{03_method}

\input{04_experiments}

\input{10_conclusion}
%
%
\bibliographystyle{splncs04}
\bibliography{main}
\end{document}

%% file: 00_abstract.tex
\begin{abstract}

We introduce a novel text-to-pose video editing method, ReimaginedAct. While existing video editing tasks are limited to changes in attributes, backgrounds, and styles, our method aims to predict open-ended human action changes in video. Moreover, our method can accept not only direct instructional text prompts but also `what if' questions to predict possible action changes. ReimaginedAct comprises video understanding, reasoning, and editing modules.  First, an LLM is utilized initially to obtain a plausible answer for the instruction or question, which is then used for (1) prompting Grounded-SAM to produce bounding boxes of relevant individuals and (2) retrieving a set of pose videos that we have collected for editing human actions. The retrieved pose videos and the detected individuals are then utilized to alter the poses extracted from the original video. We also employ a timestep blending module to ensure the edited video retains its original content except where necessary modifications are needed. To facilitate research in text-to-pose video editing, we introduce a new evaluation dataset, WhatifVideo-1.0. This dataset includes videos of different scenarios spanning a range of difficulty levels, along with questions and text prompts. Experimental results demonstrate that existing video editing methods struggle with human action editing, while our approach can achieve effective action editing and even imaginary editing from counterfactual questions. 

\end{abstract}

%% file: 01_introduction.tex
\section{Introduction\label{sec:intro}}

\begin{figure}[t]
    \centering
    \includegraphics[width=0.9\textwidth]{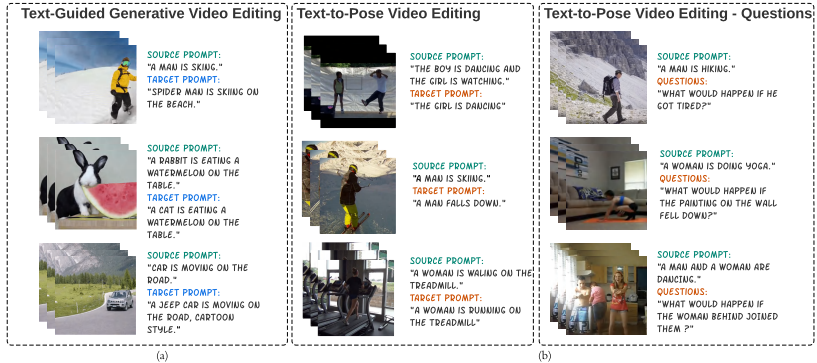}
    \caption{\textbf{Conventional Video editing and Text-to-pose video editing:} (a) Conventional Video editing directly edits a video \cite{wu2023tune, qi2023fatezero}. Similarly, Text-guided Video Editing uses a text prompt to edit the video's objects, background, style, or other attributes.  (b) \emph{Text-to-pose video editing:} directly manipulating \emph{human actions} in videos using target prompts.  \emph{Text-to-pose video editing - Question:} a what-if question is asked that would dictate the necessary modifications to support the answer to the question. This is much more challenging than video editing as the question is potentially open-ended and calls for video editing capable of accomplishing any required modifications.  \label{fig:teaser}}
\end{figure}

\begin{figure*}[t]
    \centering
    \includegraphics[width=1.0\textwidth]{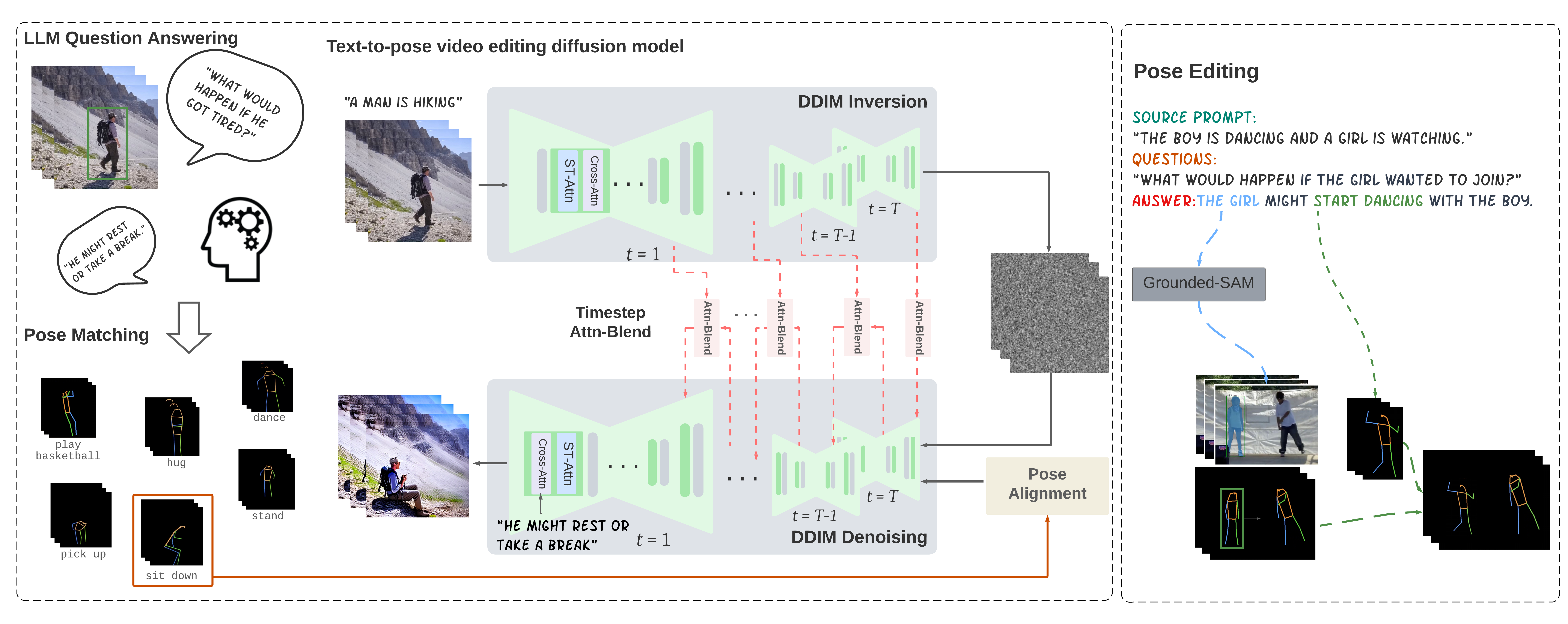}
    \caption{\textbf{Overview of \ourmethod{}:} Given a video and a  question/instruction, \ourmethod{} first uses an LLM to obtain an answer. Using the answer as a query, we conduct pose matching, after which the retrieved pose is first aligned and merged with the original pose before being used to condition our diffusion model. To handle scenarios where there could be one or more individuals, \ourmethod{} also contains a pose editing module. The pose editing module runs a Grounded-SAM model that can disambiguate the individuals needing modification based on the LLM's response. \label{fig: overview}}
\end{figure*}

\begin{figure}[t]
    \centering
    \includegraphics[scale=0.1]{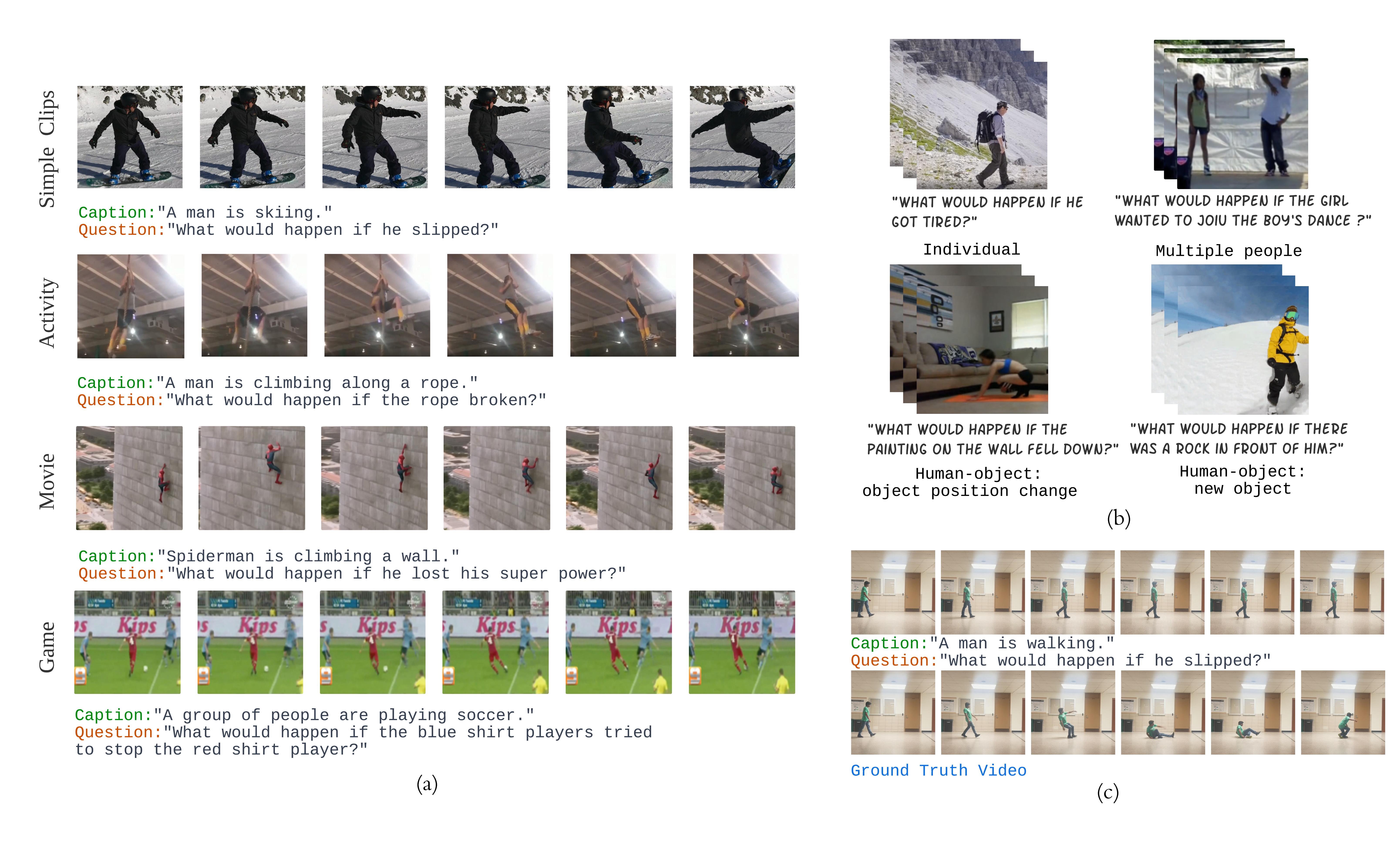}
    \caption{\textbf{Overview of \ourdataset{} Dataset:} (a) Different video categories. (b) Different scenarios with a single or multiple people and with human-object interactions. (c) Part of the dataset includes recorded videos with original videos, counterfactual questions, and associated ground truth counterfactual videos.\label{fig: dataset}}
\end{figure}

\begin{figure}[t]
    \centering
    \includegraphics[scale=0.1]{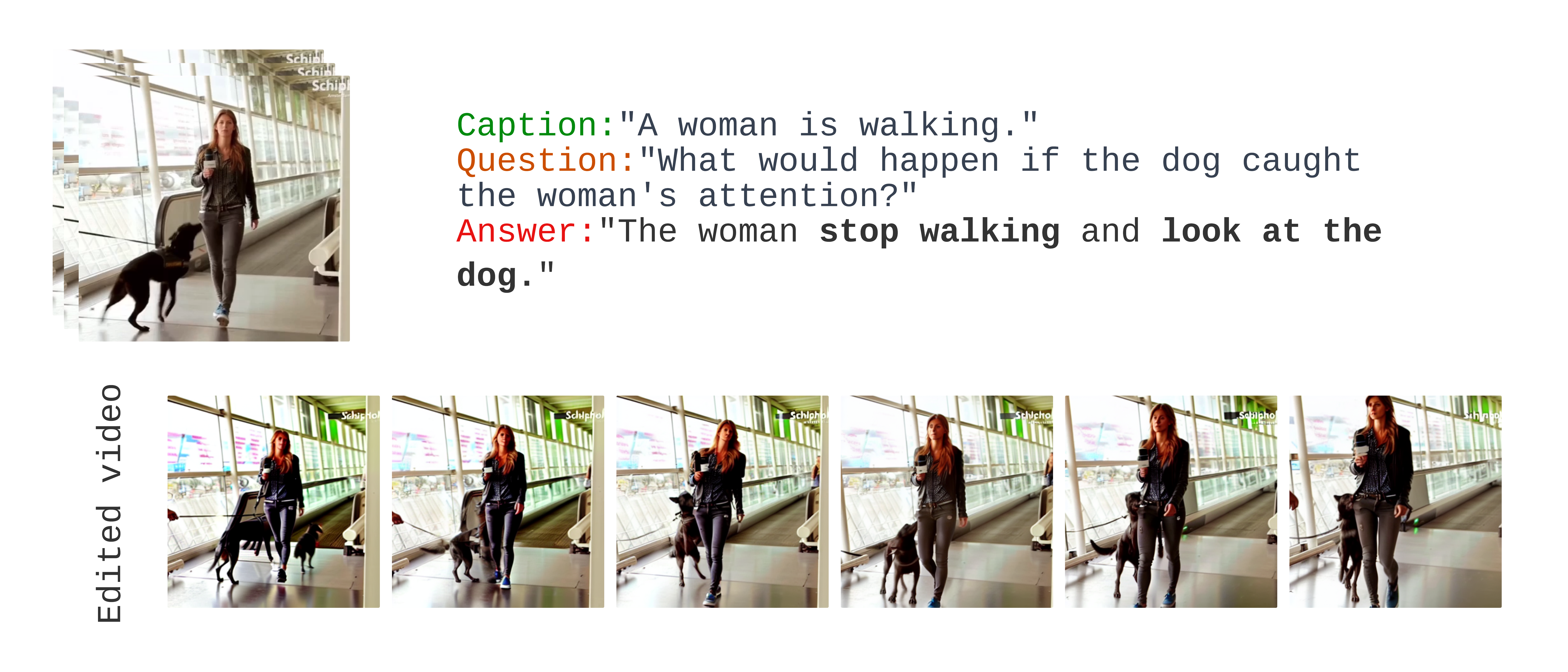}
    \caption{
    An example of a failure case. The top row is the input video, while the second is the output video. The LLM's answer contains two actions: ``stop walking" and ``look". When retrieving from the pose database, only the first action was retrieved, resulting in the final video showing the woman stopping but not looking at the dog. Our current version of \ourmethod{} contains pose videos that combine multiple actions, but not ``stop walking and look". Having all possible permutations of poses in the database is also intractable. We will leave this shortcoming of \ourmethod{} to future work.
    \label{fig: failure}}

\end{figure}

Large-scale diffusion-based text-to-video editing models are designed to generate and edit videos based on textual descriptions, capable of altering attributes, styles, and backgrounds \cite{qi2023fatezero, wu2023tune, ho2022imagen, xing2023make}. However, these editing models still struggle to manipulate human actions. Additionally, most existing methods require a reference video or other additional conditions and support only a limited variety of actions \cite{wang2023motionctrl, zhang2023controlvideo, wang2023disco}, such as dance and movement.

In contrast to the foregoing, we seek a solution that is not solely aimed at manipulating human actions but can also accept any text input - making our problem very open-ended - such as a direct instruction, a question, or even a counterfactual question, as shown in Figure \ref{fig:teaser}.
To achieve a high-fidelity solution, we first look towards the tremendous progress that has been made in the field of text-to-image generation~\cite{avrahami2022blended, couairon2023diffedit,hertz2022prompt, ruiz2022dreambooth, zhang2023adding}.  
These advancements have ushered image editing into a new era: using text-to-image diffusion models, astonishingly high-quality image editing can be achieved with simple textual descriptions. 
Unfortunately, despite the impressive progress made on the image side, high-performing video editing has remained elusive. This is because text-to-video diffusion models are limited by training costs, and not many effective models are currently available. While cost-effective approaches such as \cite{wu2023tune,wang2023zero, khachatryan2023text2video, qi2023fatezero} have been proposed, the quality of the generated videos remain low and unsuitable for action editing.

This paper introduces an effective approach called \ourmethod{} for addressing various challenges in \emph{editing human actions}. 
In \ourmethod{}, an LLM \cite{brown2020language, touvron2023llama} is first prompted to obtain an initial answer to the instruction/  question. This answer is matched with the labels of an action pose dataset that we collected and annotated. The retrieved pose video is then used to align/merge with the instance-level poses extracted from the source video with MMPose~\cite{mmpose2020}. To ensure the right individuals are modified when there are multiple people, we also employ Grounded-Segment-Anything (Grounded-SAM) \cite{kirillov2023segany,liu2023grounding} to get the segmentation masks and bounding boxes of all the individuals that need to be edited based on the LLM's answer. Together with the poses extracted from the source video, this tells us who needs to be aligned with the retrieved pose video. The aligned poses are then utilized to condition a cost-effective text-to-video diffusion model known as Tune-A-Video \cite{wu2023tune} for generating the final video. As a result, we also inherit the light fine-tuning procedure from Tune-A-Video, where only the source video is being tuned rapidly, and only the attention layers are updated. To ensure consistency between the source and generated video, we design a timestep attention blending between the foreground cross-attention maps from both the inversion and denoising process. 

To evaluate the open-ended action editing task, we propose a new dataset, \ourdataset{}, along with a benchmark. This dataset includes scenarios ranging from simple to complex and from individual to multiple people. Some example videos are shown in Figure \ref{fig: dataset}. The proposed metrics are designed to measure the temporal consistency of the generated videos, score the generated actions, and ensure that the generated videos are similar to the ground truth for those scenarios where the ground truth video is available.

While our proposed solution achieved superior results on \ourdataset{} benchmarks, we acknowledge that this paper is just a first foray into open-ended action editing. Our method may be insufficient when the questions call for sophisticated editing. For example, if the LLM's answer involves multiple actions in a sequence (e.g., ``the players jump on the ball and try to grab it''), the pose retrieval will likely only return the first action. See Figure~\ref{fig: failure} for an example. As a result, our method may also not be able to support different actions for different individuals, depending on the granularity of the answers we obtain from the LLM. Nevertheless, this paper's contributions are: 
\begin{enumerate}
    \item We introduce the concept of open-ended text-to-pose video editing, a significant and challenging task in the domain of video editing.
    \item We release a comprehensive video editing evaluation\footnote{The annotations are therefore intended for evaluations and not for training} dataset, \ourdataset{}, with accompanying measurement metrics. This dataset spans a wide variety of scenarios and difficulty levels, enabling researchers to effectively evaluate the performance and robustness of their solutions. 
    \item We introduce a novel method, \ourmethod{}, that serves as a strong baseline for future work. Our method allows for changes in the  background, objects, style, events that have already occurred in the video, and alterations in the actions and positions. As part of our method, we also collected a comprehensive action pose dataset, which we will release to guide human action editing.
\end{enumerate}

%% file: 02_relatedwork.tex
\section{Related Work}
\label{sec:relatedwork}

\noindent
\textbf{Text-guided Video Generation.} Text-to-Video Generation translates textual descriptions into coherent videos by visualizing the scenes, objects, and actions from the text \cite{xing2023survey}. 
Drawing inspiration from Text-to-Image diffusion models, Video Diffusion Models \cite{ho2022video} extends image diffusion models to video generation by employing a 3D U-Net architecture and joint image-video training. Make-A-Video \cite{singer2022make} improves visual quality by training spatial super-resolution and frame interpolation models, boosting video resolution and achieving a higher, adjustable frame rate. Imagen Video \cite{ho2022imagen} utilizes cascaded video diffusion models to generate high-definition and high-fidelity videos. 

Although T2V generative models achieve impressive results, they demand extensive training data and significant computational resources. Therefore, some approaches switch their attention to zero-shot techniques for T2V generation. Text2Video-Zero~\cite{khachatryan2023text2video} introduces zero-shot text-to-video synthesis by modifying text-to-image models with motion-enriched latent codes and cross-frame attention, enabling low-cost, high-quality video generation without training, and extending to various video generation tasks. Free-bloom \cite{huang2023free} leverages large language models for semantic prompt creation and latent diffusion models for frame synthesis, ensuring semantic, temporal, and identical coherence through novel annotative modifications. FlowZero \cite{lu2023flowzero} uses dynamic scene syntaxes to guide the generation of temporally-coherent videos from text, enhancing frame-to-frame coherence and global alignment through iterative self-refinement and motion-guided noise shifting.

\noindent
\textbf{Text-guided Video Editing.} Since video diffusion models (VDM) such as \cite{ho2022video, ho2022imagen} were introduced, researchers have started leveraging these powerful models for video editing tasks. Dreamix~\cite{molad2023dreamix} performs mixed video-image fine-tuning on VDM, enabling edits to the video's objects and even motions. Gen-1~\cite{esser2023structure} introduced a content-aware structure model, allowing video edits based on reference images or textual descriptions. These methods rely heavily on pretrained VDMs, which are not publicly available, and also require significant computational resources for training on large-scale datasets. As a result, researchers have sought more computation-friendly approaches. Tune-A-Video~\cite{wu2023tune} inflates the existing image diffusion model to 3D, after which fine-tuning is conducted on the source video, achieving effective video editing. FateZero~\cite{qi2023fatezero} captures intermediate attention maps during inversion that preserve structural and motion details, subsequently utilizing them for zero-shot editing of style and object attributes. Pix2Video \cite{ceylan2023pix2video} proposed to use self-attention features for progressive propagation without the need for fine-tuning on specific videos. In order to maintain the context, appearance, and identity of the foreground object, Ground-A-Video\cite{jeong2023ground} proposes Cross-Frame Gated Attention and Modulated Cross-Attention with optical flow smoothing to ensure temporal consistency and edit accuracy, achieving training-free approach for multi-attribute video editing.

Video editing models also accept other conditions such as pose, motion, and user guidance. As a training-free framework adapted from ControlNet, \cite{zhang2023controlvideo} enhances text-to-video generation with structural consistency, incorporating cross-frame self-attention, an interleaved-frame smoother for flicker reduction, and a hierarchical sampler for efficient long video synthesis, achieving generating coherent videos in commodity GPUs. DragVideo \cite{deng2023dragvideo} proposes a drag-style video editing framework which enables intuitive user control, incorporating Sample-specific LoRA and Mutual Self-Attention to ensure temporal consistency results across diverse editing tasks. Leveraging a pre-trained LVDM model, MotionCtrl \cite{wang2023motionctrl} maintains video integrity while offering a broad range of motion customizations.

While most existing methods rely on additional condition information or reference videos, our method allows for greater flexibility with text prompts, including editing actions in imagined scenes, without using control information.

%% file: 03_method.tex
\section{Text-to-pose Video Editing\label{sec:task}}

\subsection{Problem Definition\label{sec:problem_def}}
Given a source video  $\mathcal{V}$ and a text/ question, $\mathcal{Q}$, the task is to predict specific attributes of $\mathcal{V}$ that needs to be altered to effectively answer $\mathcal{Q}$. The required content change and consequently the output video $\mathcal{V'}$ are then to be generated with high fidelity,
while ensuring that $\mathcal{V'}$ stays faithful to $\mathcal{V}$.

\subsection{\ourdataset{} Dataset \label{sec:dataset}}

To enable the evaluation of different models, we proposed the \ourdataset{} dataset, which contains 101 videos of different scenes with captions and questions. Presently, practitioners should only utilize \ourdataset{} for evaluation purposes and avoid including them in the training. We plan to release training data in future versions of \ourdataset{}. The dataset comprises five different scenarios: \emph{simple clips}, 
\emph{human activities}, \emph{highlight clips from movies}, \emph{highlight clips from sports}, \emph{daily scenes recorded by authors}. In each scenario, there are three levels of difficulty (easy, medium, hard) and three different types of interactions (individual, multiple people, and human-object interaction). For \emph{daily scenes}, we further recorded ``pseudo'' ground truth videos. For each video, we annotated a caption and designed a counterfactual question with assistance from ChatGPT4 \cite{OpenAI2023GPT4TR}.

\noindent\textbf{Simple clips} contain videos that are frequently used for evaluating video editing models. These videos are from the DAVIS dataset \cite{pont20172017}. They correspond to simple scenarios, largely involving a single individual or scenarios with simple human-object interactions, from a fixed camera perspective or with limited motion.

\noindent\textbf{Human activity} clips are videos selected from the existing CausalVideoQA dataset~\cite{li2022from}. It is a VideoQA dataset that includes a large number of videos and questions. However, not all videos can be directly utilized for the text-to-pose video editing task since the scenes are either too challenging for video editing, or the questions are too open-ended. So we manually selected usable videos and categorized them by difficulty levels.

\noindent\textbf{Movie highlights} includes some classic movie clips. The ability to generate a new version of a movie with counterfactually motivated variations of the script has great utility in real-world applications. For instance, a small factual change in the script could lead to an entirely different storyline over a long horizon. We believe these clips will be useful to the community for evaluating the effectiveness and practical utility of a given approach for text-to-pose video editing.

\noindent\textbf{Sports clips} includes videos from different competitive sports. Their questions relate to the changes in the competitions due to alterations in the players' actions or unexpected external factors.

\noindent\textbf{Daily scenes.} We further recorded an additional set of videos containing not only the source video but also the corresponding ground truth videos. These videos were performed by different actors and span a variety of settings, including both indoor and outdoor scenes such as offices, classrooms, libraries, basketball courts, tennis courts, lawns, and walking trails. They featured single and multiple individuals acting out different daily scenarios. The ground truth video is acted out by the actors carefully maintaining the same settings, as much as possible, except where modifications in response to the question are necessary. In this sense, the ground-truth videos could be considered ``pseudo,'' but close examination will reveal their high quality. To ensure the validity of the ground truth videos, actions were based on answers solicited from an LLM by prompting it with the question. In cases where the LLM provided multiple answers, corresponding ground truth videos were also recorded.

\section{Method\label{sec:method}}

In Sec.~\ref{sec:diffusion_inversion}, we first introduce the latent diffusion model, DDIM  and Tune-A-Video~\cite{wu2023tune} diffusion engine that generates the counterfactual video.
In Sec.~\ref{sec:cv_editing}, we first describe how we would prompt the LLM to get proper answers to questions. Following that, we present how we modify the poses of individuals to meet the required changes, which is then used to condition the diffusion model. 

\subsection{Diffusion Model \label{sec:diffusion_inversion}}

\noindent\textbf{Latent Diffusion Models (LDM)} \cite{rombach2022high} works in the latent space of powerful pretrained autoencoders. Encoder $\mathcal{E}$ compresses images $x$ to latent $z=\mathcal{E}(x)$, and Decoder $\mathcal{D}$ reconstructs it back to image $ \mathcal{D}(z) \approx x $.
A time-conditional UNet  $\varepsilon_\theta (o,t)$ 
is trained to remove the noise using the objective:

\begin{equation}
L_{LDM}:= \mathbb{E}_{\mathcal{E}(x),\varepsilon_\theta \sim \mathcal{N}(0,1),t}[||\varepsilon-\varepsilon_\theta(z_t,t)||^2_2].
\end{equation}

\noindent\textbf{DDIM} \cite{song2020denoising} is a deterministic sampling process, which converts a random noise to clean latent in the  steps  $t: T\rightarrow 1$:

\begin{equation}
    z_{t-1} = \sqrt{\alpha_{t-1}} \; \frac{z_t - \sqrt{1-\alpha_t}{\varepsilon_\theta}}{\sqrt{\alpha_t}}+ \sqrt{1-\alpha_{t-1}}{\varepsilon_\theta},
\end{equation}
where $\alpha_{t}$ is a parameter for noise scheduling. 
Stable Diffusion guides the diffusion processes with a textual prompt $\tau$ for text-to-image generation:
\begin{equation}
\begin{split}
    z_{t-1} & = \sqrt{\alpha_{t-1}} \; \frac{z_t - \sqrt{1-\alpha_t}{\varepsilon_\theta}(z_t,t,\tau)}{\sqrt{\alpha_t}} \\ &+ \sqrt{1-\alpha_{t-1}}{\varepsilon_\theta}(z_t,t,\tau),
\end{split}
\end{equation}

\noindent\textbf{Inflated Diffusion Model.} While LDM is  designed for 2D input, we use the  spatio-temporal  structure of LDM from Tune-A-Video \cite{wu2023tune}. The 2D convolution layers are inflated to pseudo-3D convolution layers by replacing $3 \times 3$ kernels with $1 \times 3 \times 3$.
The spatial self-attention is further extended to spatio-temporal attention to model temporal consistency.
Additionally, \cite{wu2023tune} also used sparse causal attention (SCA), which limits the model to compute the attention matrix using only the first and previous frames to reduce the computational cost. By tuning those attention layers, \cite{wu2023tune} is able to generate temporally consistent videos by using prior knowledge from pretrained Text-to-Image models.

\subsection{Text-to-pose Video Editing\label{sec:cv_editing}}

As shown in Figure \ref{fig: overview}, \ourmethod{} first uses an LLM to obtain an answer and reasoning to the given questions. A Grounded-SAM model is applied for video understanding to identify the individuals to be edited. Then, pose matching and alignment are utilized to guide the action edit in video. Finally, a inflated T2I diffusion model with a novel timestep attention blending module is designed to achieve high-quality human action editing in videos. 

\begin{figure}[h]
    \centering
    \includegraphics[width=0.4\textwidth]{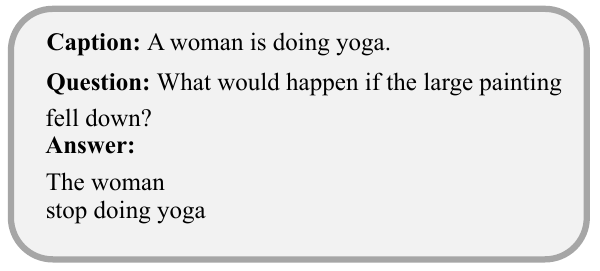}
    \caption{Example prompting template.}
    \label{fig:enter-label}
\end{figure}

\noindent\textbf{Question Answering.} Given the questions, we first use GPT4 to generate an answer. We use a question template comprising the caption and the prompt ``what would happen if...", where the caption serves as context for the LLM. However, as expected, the answers from LLM tend to contain redundancy and verbiage. We mitigate this by listing some template questions and answers to instruct LLM on how to respond to new questions. Figure~\ref{fig:enter-label} shows an example template.

\noindent With this, the LLM started generating answers in the form of ``someone" ``do something", which helps the model understands the person that needs to be edited.

\noindent\textbf{Pose Matching.} 
Text-to-pose video editing entails modifications in individual's  actions, positions. Therefore, the ability to effect changes in people's actions and position is an essential step. We consider using pose videos as a condition to guide these changes. Specifically, we draw inspiration from Followyourpose \cite{ma2023follow}  model, which introduced pose videos as an additional input as control information. Given a action-related pose video, we combine the feature maps of the pose video with that of the original video to effectively guide the diffusion model to change people's actions. We retrieve the most suitable pose video by conducting  a similarity search in an action pose dataset, using the Bert language model \cite{reimers-2019-sentence-bert} to compute the similarity scores between LLM's answers and all pose labels. This action pose dataset is collected from a large number of classic action-related videos from action recognition datasets and YouTube videos, covering various scenarios and common behaviors. We use mmpose \cite{mmpose2020} to extract pose videos, and label each with its corresponding action description, like ``dance", ``sit down", ``fall down", etc. Since the matched pose video may have a significant geometrical delta from the original pose, we perform scaling, translation and rotation alignment by Procrustes analysis:

\begin{equation}
    \min_{\mathbf{s},\mathbf{R},\mathbf{t}}||X_{ps}-(\mathbf{s}\mathbf{R}X_{pd}+\mathbf{t})||^2_2,
\end{equation}
where $X_{ps}$ are the keypoints for the first frame of source video and $X_{pd}$ are the keypoints for the first frame of matched pose video. After obtaining scaling $\mathbf{s}$, translation $\mathbf{t}$, and rotation $\mathbf{R}$, we applied them to all frames of the pose video.

\noindent\textbf{Pose editing module.}  \ourmethod{} is designed to handle scenarios with one or more people in the video. In order to identify the right individuals to be modified, we first prompt the Grounded-SAM~\cite{kirillov2023segany,liu2023grounding} with the answer to detect them, obtaining their segmentation masks
and bounding boxes. Then we perform pose estimation \cite{mmpose2020} on the source video and alter poses of individuals that have been detected by Grounded-SAM to the retrieved poses.
As an example, in Figure \ref{fig: overview} (b), the individual to be edited is ``the girl". The Grounded-SAM first 
detect the segmentation mask and bounding box of ``the girl". Then, in the pose video, the standing pose of the girl is replaced and aligned with the 
retrieved ``dance" pose. Finally, we obtained an edited pose video where both the girl and the boy are dancing.

\noindent\textbf{Text-to-pose diffusion model.} 
Our text-to-pose diffusion model adopts the inflated text-to-image model \cite{wu2023tune} or what is commonly known as Tune-A-Video as base model, using the edited pose video as its condition. Following \cite{ma2023follow}, we inject the pose guidance into the U-Net structure of the base model via a residual connection that allows us to add the pose feature into each layer. 
To help the model correctly detects regions that need to be changed while staying faithful to the source, we conduct attention blending between the inversion and denoising process. However, standard attention blending \cite{qi2023fatezero} is not suitable for action editing because their foreground to be edited is always in the same location as source videos, and there was no alteration in people's action. Since our algorithm involves changes to the action, which may also lead to changes in the positions of individuals, a fixed attention mask generated from the inversion process is not applicable. Therefore, we propose a timestep attention blending. During the editing process, we first obtain an initial mask $M^T$ by thresholding the cross-attention maps from the inversion process $c_{inv}^T$, and blend the self-attention map of the inversion $s_{inv}^T$ and the denoising $s_{den}^T$ of the $T-th$ step by the mask,
 \begin{equation}
    s_{edit}^T = M^T\odot s_{den}^T + (1-M^T)\odot s_{inv}^T.
\end{equation}
Then for the rest of the time steps, we use the cross-attention maps of denoisng process $c_{den}^t$ from the last time step to threshold the mask $M^t$ so that it also includes the area where action changes happen,
\begin{equation}
    s_{edit}^{t-1} = M^t\odot s_{den}^{t-1} + (1-M^t)\odot s_{inv}^{t-1},
\end{equation}
where $t: T\rightarrow 2$. With the help of cross-attention maps from the denoisng process, the mask not only covers the foreground from the source video, but also potential foreground in the edited video, which effectively indicates the area to be edited while maintaining the consistency of the other parts.

\begin{figure}[]
    \centering
    \includegraphics[width= \textwidth]{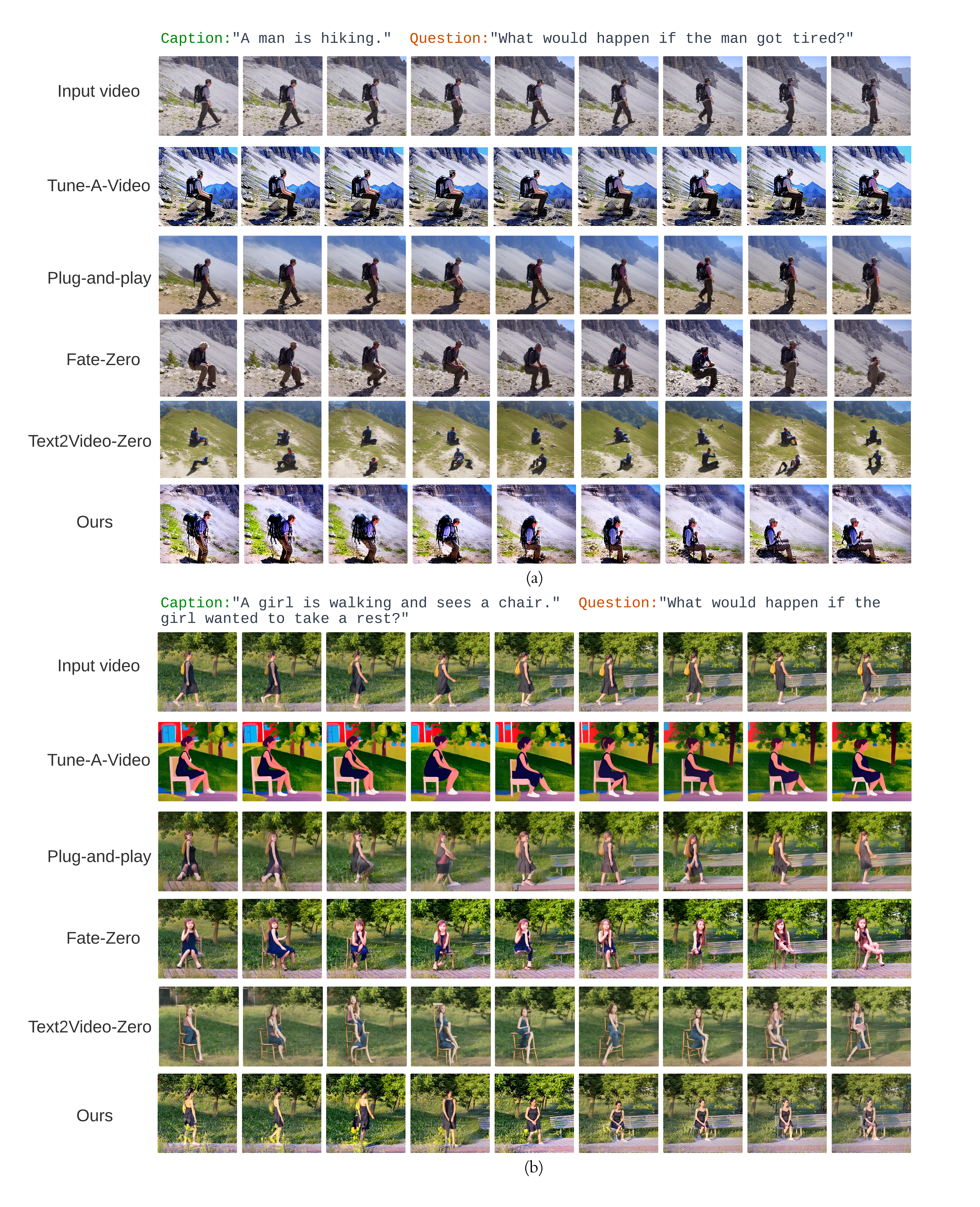}
    \caption{Comparing \ourmethod{} with baselines qualitatively. (a)  Other baselines failed to generate the full range of the action of sitting down to rest. (b) Tune-A-Video generates a video with style changes that are inconsistent with the original video, while other baselines failed to generate the full range of the action of sitting. \label{fig:qualitative}}
\end{figure}

%% file: 04_experiments.tex
\section{Experiments\label{sec:experiments}}

\subsection{Implementation Detail}

We develop our model based on Stable Diffusion and the pre-trained weights from \cite{ma2023follow}. Following \cite{wu2023tune}, the source video is fine-tuned once to retain both the foreground and context information. In the fine-tuning stage, we sample 12 frames from the source video uniformly and fine-tune the models with our method for 300 steps with a learning rate of $3 \times 10^{-5}$ and a batch size of 1. At inference, we generate 12 frames for each video in normal setting, and 24 frames for each video in long video editing setting. The resolution is $512 \times 512$ for both fine-tuning and inference.

\subsection{Baseline Comparisons}

\noindent

\noindent
\textbf{Baselines.}
 We compare with Tune-A-Video \cite{wu2023tune}, Plug-and-play \cite{tumanyan2023plug}, Fate-Zero \cite{qi2023fatezero}, Text2Video-Zero \cite{khachatryan2023text2video} which accept arbitrary text inputs for video editing.

\noindent
\textbf{Quantitative results.}
We conduct a quantitative evaluation using pretrained VideoCLIP \cite{xu-etal-2021-videoclip} and CLIP\cite{radford2021learning} to produce three metrics: (1) \textbf{Vid-Acc} is the video-wise editing accuracy, which is the percentage of edited videos with a higher VideoCLIP similarity to the target prompt (LLM answer) than the source prompt (video caption). (2) \textbf{Vid-Con} measures the frame-wise consistency based on cosine similarity between the CLIP  features of the edited videos and source video. It evaluates if the edited video is faithful to the source video and not just some randomly generated imaginary videos from text prompts.
(3) Since the Daily set has ground truth video, we propose another metric, \textbf{GT-Con}, which computes the cosine similarity between the CLIP features of the edited video and ground truth video. Based on these three metrics, both Table \ref{tab:quantitative} and \ref{tab:quantitative_gtcon} demonstrate the superior performance of \ourmethod{}, which outperform other baselines in most scenarios and also has the highest average Vid-Acc and Vid-Con. We do however observe that in Table \ref{tab:quantitative}, Fate-zero has good Vid-Con, because the method preserves the background well, which can also be seen from the Figure \ref{fig:qualitative}. However, the method is unable to achieve good video editing. In the Activity set, Tune-A-Video
better Vid-Acc because their generated videos match the text prompt well. That said, the videos were very different from the original videos, which can be seen from their low Vid-Con, which highlights the importance of reading the performance numbers in totality. Similarly, as can be seen from Table \ref{tab:quantitative_gtcon}, \ourmethod{} also achieves better results in the presence of ground truth videos. This indicates that merely maintaining background consistency is not sufficient. The method needs to achieve correct action editing to better align with the ground truth videos.

\begin{figure}[t]
    \centering
    \includegraphics[width=0.9\textwidth]{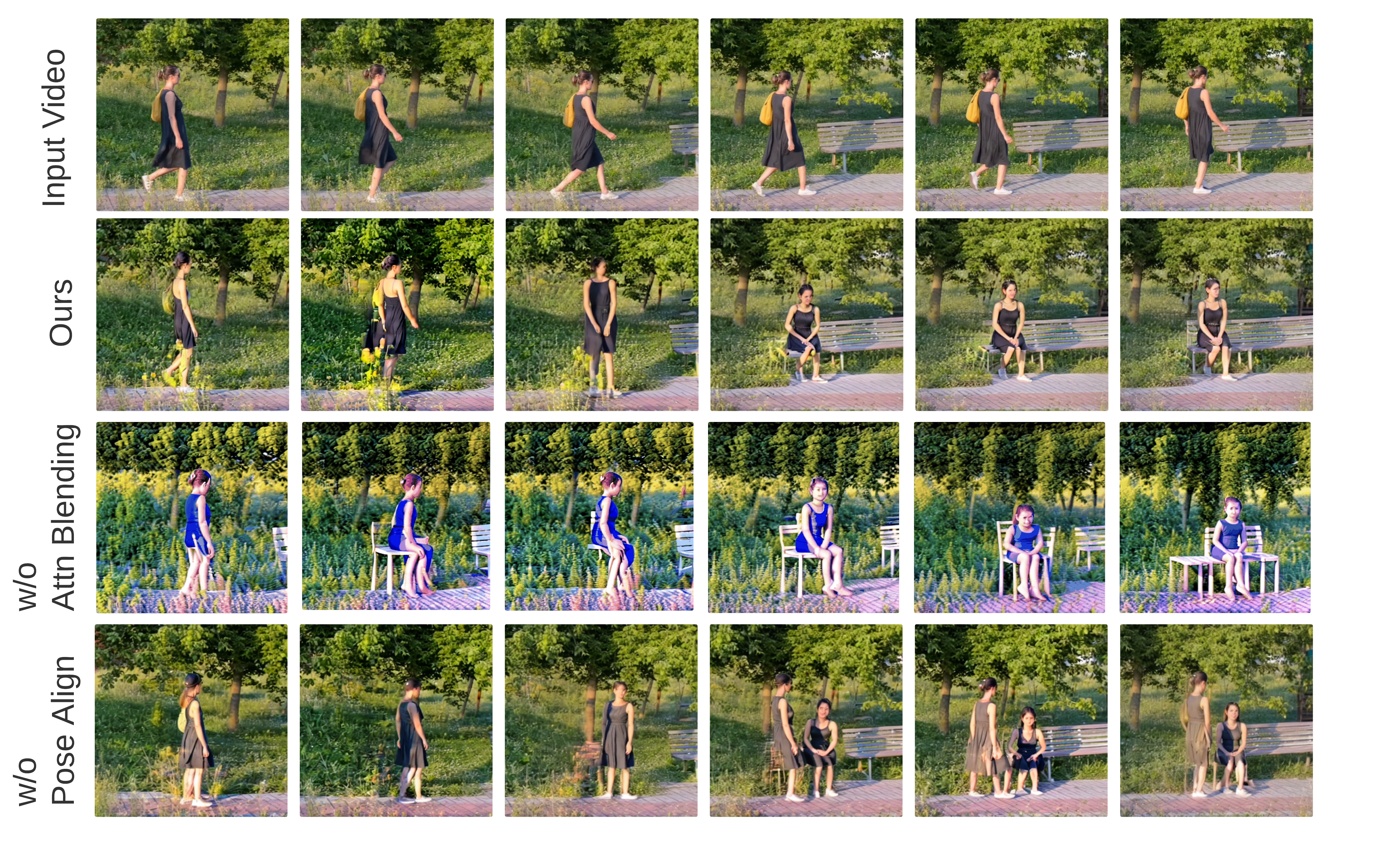}
    \caption{Ablation Study on (a) timestep attention blending module, and (b) pose alignment module. More can be found in the supplementary.}
    \label{fig:ablation_res}
\end{figure}

\begin{figure}[t]
    \centering
    \includegraphics[width=0.9\textwidth]{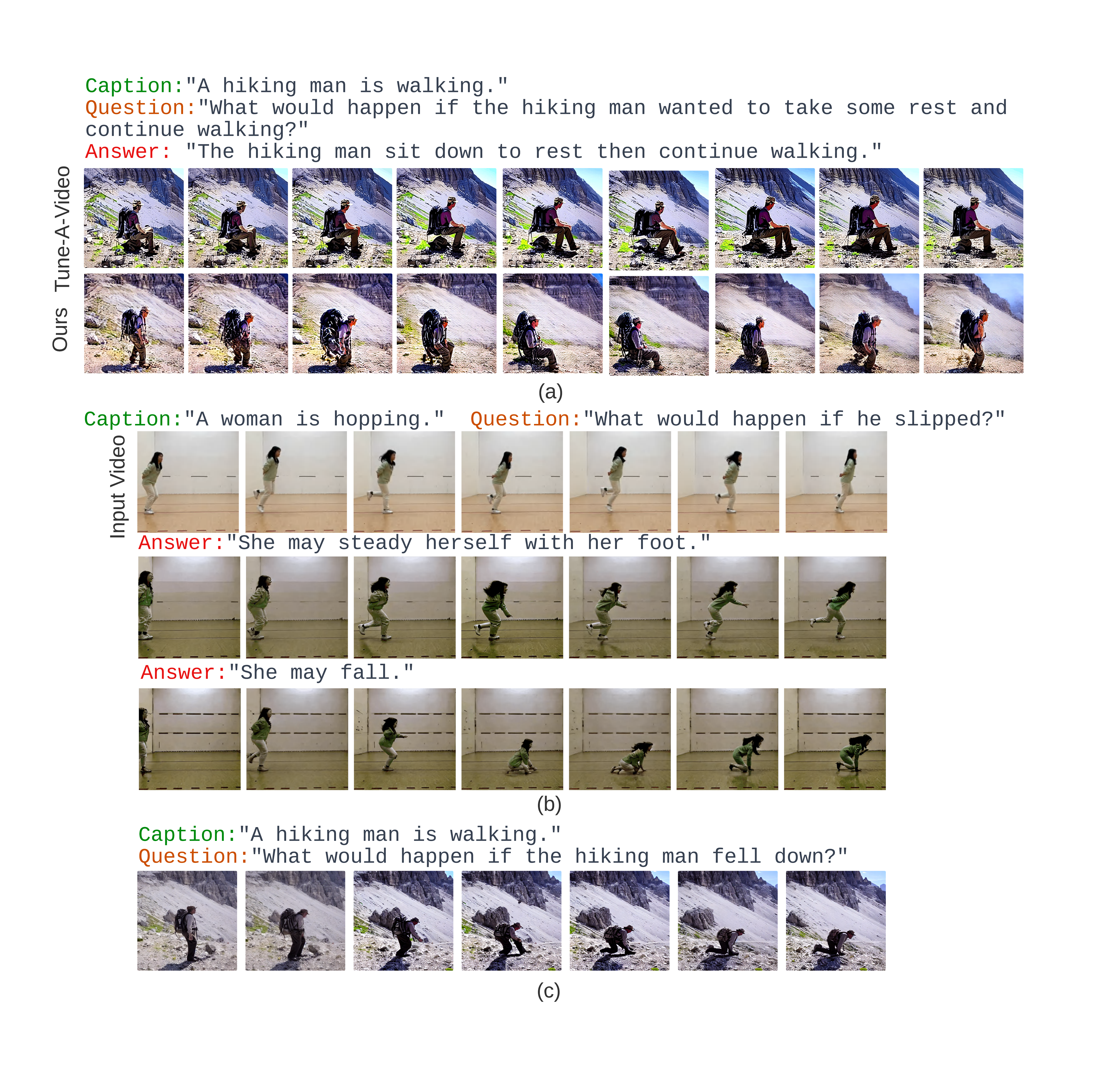}
    \caption{(a) Long video generation results. Input video is in Figure \ref{fig:qualitative}. (b) and (c) Diverse results.}
    \label{fig:longvideo_diverse}
\end{figure}

\noindent
\textbf{Qualitative results.} In Figure \ref{fig:qualitative}, we show some of the strengths of \ourmethod{} in regards to the action editing task. In general, \ourmethod{} appears to generate higher fidelity action changes. We also observe that \ourmethod{} can stay faithful to the original video, especially in terms of style and background.

\noindent
\textbf{Long video results.} We also look at creating longer frames video with continuous action changes. As shown in Figure~\ref{fig:longvideo_diverse} (a), after the hiker took a rest, according to the answer, the hiker will stand up and continue walking. Here, we observe that Tune-A-Video can only generate an action combining ``sit" and ``walk", and it seems unable to generate the full progression of the action. In contrast, our method accurately predicts the entire process from the hiker sitting down to rest to standing up and continuing hiking.

\noindent
\textbf{Diverse results.} We study the ability of \ourmethod{} to generate diverse results. When LLM provides multiple answers, \ourmethod{} is able to generate a new outcome for each answer with high fidelity as shown in the Figure ~\ref{fig:longvideo_diverse} (b). Given any question or instruction, our method can generate corresponding videos with changes in action. For example, Figure ~\ref{fig:longvideo_diverse} (c) shows a scenario where we want the hiking man in the video to fall. The result demonstrates that our questions or instructions are open-ended, and they are not limited by the dataset, showing high freedom of action editing.

\noindent
\textbf{Ablation Study.}
Figure ~\ref{fig:ablation_res}  shows the impact of different modules. The first and third row show the result before and after removing the attention blending module. We observe that both the foreground and background become significantly different from the source video. This result highlights the importance of the role the attention module play in preserving identity and maintaining background consistency. The last row shows the before and after removing the pose alignment module. Without pose alignment, we observe that the girl become smaller, especially when she sits. In some frames, there is a ``ghost" shape of the girl from the source video. This is because when the pose is not well aligned, the attention mask tend to fail to localize the region to be edited correctly. 
Similarly, it can be observed from Table \ref{tab:quantitative_gtcon} that after the absence of these two important modules, the GT-Con decreases. We note that the attention blending module has a greater impact on GT-Con because in most videos the background occupies a large part spatially.

\begin{table}[t] \tiny
\centering
\caption{Quantitative comparison with Baselines on \ourdataset{}.}
\label{tab:quantitative}
\begin{tabular}{lllllllllll}
\hline
\multirow{2}{*}{Method} & \multicolumn{2}{c}{Simple Clips}                              & \multicolumn{2}{c}{Activity}                                  & \multicolumn{2}{c}{Movie}                                     & \multicolumn{2}{c}{Game} 
& \multicolumn{2}{c}{Average}      \\ \cline{2-11} 
                        & \multicolumn{1}{c}{Vid-Con} & \multicolumn{1}{c}{Vid-Acc} & \multicolumn{1}{c}{Vid-Con} & \multicolumn{1}{c}{Vid-Acc} & \multicolumn{1}{c}{Vid-Con} & \multicolumn{1}{c}{Vid-Acc} & \multicolumn{1}{c}{Vid-Con} & \multicolumn{1}{c}{Vid-Acc} & \multicolumn{1}{c}{Vid-Con}
                        & \multicolumn{1}{c}{Vid-Acc}\\ \hline
Tune-A-Video \cite{wu2023tune}           &    0.696                           &       0.384                        &             0.725                  &        \textbf{0.470}                       &        0.804                       &          0.235                     &                         0.715      &      0.250       &      0.739    &    0.338      \\ 
Plug-and-Play \cite{tumanyan2023plug}          &   0.797                            &       0.384                        &      0.790                         &      \underline{0.411}                         &          0.829
&         0.294                      &           0.735                    &     \textbf{0.416}        &      0.791      &    0.372    \\ 
Fate-Zero \cite{qi2023fatezero}    &     \textbf{0.868 }      &       0.230                        &       0.903                       &    0.352                            &      \textbf{0.915 }                        &          \textbf{0.411}                     &           0.882                    &             \underline{0.333}          &      0.894    &    0.372                           \\ 
Text2Video-Zero  \cite{khachatryan2023text2video}             &    0.715                           &         \textbf{0.461}                      &    0.651                           &       0.352                        &          0.672                     &     0.294                          &       0.728                        &     0.083    &  0.686   &    0.305               \\ 
Ours                  &            \underline{0.867}                     &          \textbf{0.461}                     &                    \textbf{0.917}           &          \underline{0.411}                    &            \underline{0.905}                   &       \underline{0.352}                        &    \textbf{0.902}                           &           \underline{0.333}         &   \textbf{0.899}      &  \textbf{0.389}  \\ \hline
\end{tabular}
\end{table}

\begin{table}[t]
\centering
\caption{Ablation study and Baseline comparison with ground truth based on the test split of the Daily set. 
}
\label{tab:quantitative_gtcon}
\begin{tabular}{lll}
\hline
\multicolumn{2}{c}{Method}                                   & \multicolumn{1}{c}{GT-Con} \\ \hline
\multirow{2}{*}{Ablation}  & w/o timestep attention blending &           0.848                 \\
                           & w/o pose align                  &     0.866                       \\ \hline
\multirow{4}{*}{Baselines} & Tune-A-Video  \cite{wu2023tune}                  &                0.723            \\
                           & Plug-and-Play   \cite{tumanyan2023plug}                &      0.731                      \\
                           & Fate-Zero   \cite{qi2023fatezero}                    &          0.853                  \\
                           & Text2Video-Zero \cite{khachatryan2023text2video}                &    0.715                        \\ \hline
\multicolumn{2}{l}{Ours}                                     &                  \textbf{0.869 }         \\ \hline
\end{tabular}
\end{table}

%% file: 10_conclusion.tex
\section{Conclusion}

We propose \ourmethod{} as the first baseline to predict action changes from text, questions or even counterfactual questions without relying on additional references. Different from text-guided generative video editing, which directly points out changes to the video's attributes, foreground, background, style, etc., our task requires the model to first predict the consequences of the questions, before generating the corresponding edited videos. 
\ourmethod{} includes different effective modules to achieve changes to human actions, and is also able to handle scenarios with multiple people. 
We also provided the \ourdataset{} dataset, which allows this task to be effectively evaluated by other researchers in the future.

\noindent\textbf{Limitations} More work will need to be done for \ourmethod{} to more effectively handle complex scenarios such as interactions with other objects, and handling multiple people. A comprehensive training dataset that compliments \ourdataset{} will also be an important extension of this work.